\documentclass[letterpaper, 10 pt, conference]{ieeeconf}  % Comment this line out if you need a4paper

\usepackage[noadjust]{cite}
\usepackage{amsfonts}
\usepackage{amsmath}
\usepackage{xcolor}
\usepackage{amssymb} % for \smallsetminus
\usepackage{diagbox} % for \backslashbox
\usepackage{lipsum}  % for random text, to be removed
\usepackage{multirow} % for \multirow
\IEEEoverridecommandlockouts                              % This command is only needed if 
\usepackage{graphicx}
\title{\LARGE \bf
Bathymetric Surveying with Imaging Sonar Using Neural Volume Rendering
}

\author{Yiping Xie$^{1}$, Giancarlo Troni$^{2}$, Nils Bore$^{3}$ and John Folkesson$^{1}$ % <-this % stops a space
\thanks{*This work was supported in part by the Wallenberg AI, Autonomous Systems and Software Program (WASP) funded by the Knut and Alice Wallenberg Foundation and in part by Stiftelsen för Strategisk Forskning (SSF) through
the Swedish Maritime Robotics Centre (SMaRC) under Grant IRC15-0046.
(\textit{Corresponding author: Yiping Xie.})}
\thanks{$^{1}$ Yiping Xie and John Folkesson are with Robotics Perception and Learning Division, KTH Royal Institute of Technology, Stockholm, Sweden. (e-mail: \{yipingx, johnf\}@kth.se)}%
\thanks{$^{2}$ Giancarlo Troni is with Monterey Bay Aquarium Research Institute, Moss Landing, CA, USA. (email: gtroni@mbari.org)}
\thanks{$^{3}$ Nils bore is with Ocean Infinity. (email: nils.bore@oceaninfinity.com)}
}

\begin{document}

\maketitle
\thispagestyle{empty}
\pagestyle{empty}

%%%%%%%%%%%%%%%%%%%%%%%%%%%%%%%%%%%%%%%%%%%%%%%%%%%%%%%%%%%%%%%%%%%%%%%%%%%%%%%%
\begin{abstract}
 
This research addresses the challenge of estimating bathymetry from imaging sonars where the state-of-the-art works have primarily relied on either supervised learning with ground-truth labels or surface rendering based on the Lambertian assumption.
In this letter, we propose a novel, self-supervised framework based on volume rendering for reconstructing bathymetry using forward-looking sonar (FLS) data collected during standard surveys. We represent the seafloor as a neural heightmap encapsulated with a parametric multi-resolution hash encoding scheme and model the sonar measurements with a differentiable renderer using sonar volumetric rendering employed with hierarchical sampling techniques. Additionally, we model the horizontal and vertical beam patterns and estimate them jointly with the bathymetry. We evaluate the proposed method quantitatively on simulation and field data collected by remotely operated vehicles (ROVs) during low-altitude surveys. Results show that the proposed method outperforms the current state-of-the-art approaches that use imaging sonars for seabed mapping. We also demonstrate that the proposed approach can potentially be used to increase the resolution of a low-resolution prior map with FLS data from low-altitude surveys.

\end{abstract}

%%%%%%%%%%%%%%%%%%%%%%%%%%%%%%%%%%%%%%%%%%%%%%%%%%%%%%%%%%%%%%%%%%%%%%%%%%%%%%%%
\section{INTRODUCTION}
Acquiring high-resolution bathymetry is one of the most fundamental and crucial applications to underwater exploration. Traditionally, multibeam echo sounders (MBES) are the de facto sensors for collecting bathymetric data, however, recently there has been an increasing interest in the research topic of reconstructing bathymetry using imaging sonars, e.g., forward-looking sonars (FLS). FLS sensors have significant capabilities in underwater environments with many advantages, compared to other sensors, making them suitable for various applications. Compared to side-scan sonars (SSS) and MBES, FLS measurements have overlap between consecutive frames, and its high-resolution imagery makes it ideal for low-altitude surveys. While comparing to optical cameras, FLS has longer range measurements and can operate in murky underwater environments with low-light visibility. The main problem is that though FLS (as with SSS) resolves range and azimuth angle, there is an ambiguity (especially for wide-aperture sonars) in the elevation angle in the sense that the return can come from anywhere along the elevation arc at given range and azimuth angle. 
\begin{figure}[t]
\centering
\includegraphics[width=\linewidth]{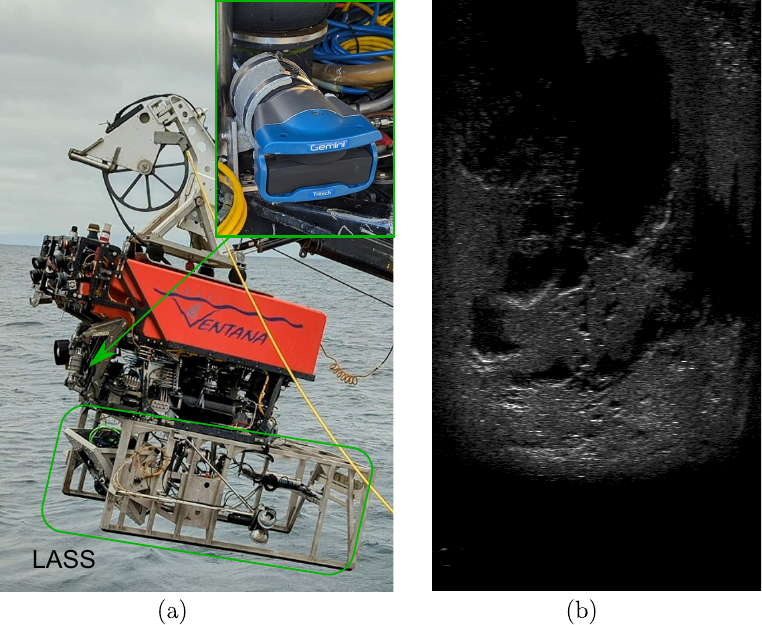}  
  \caption{(a) ROV Ventana with the Low Altitude Survey System (LASS) mounted underneath being deployed from R/V Rachel Carson. The top right corner shows the sonar to acquire the field images. (b) An example of a sonar image captured with Gemini 720is FLS.}
  \label{fig:ROV_LASS}
\end{figure}
Much research has been focusing on how to resolve the elevation ambiguity for object reconstruction, among which learning-based methods, more recently have achieved state-of-the-art results. However, existing algorithms are rarely suitable for this work's application, which is, high-resolution, dense seabed mapping.

A lot of early works have mainly focused on feature-based, sparse 3D reconstruction~\cite{huang2016incremental,mai20173,westman2018feature,li2018pose,westman2019degeneracy}, while other methods focusing on dense reconstruction suffer from various limitations, such as enforcing restrictions on elevation aperture~\cite{teixeira2016underwater,teixeira2019dense}, the motion of the vehicle~\cite{westman2020theory,wang2023motion}, relying on information from object edges/shadows~\cite{aykin20153,aykin2016modeling,westman2019wide} or relying on volumetric grids~\cite{guerneve2015underwater,guerneve2018three,wang20183d,westman2020volumetric} that are expensive when the scene is large with fine-grained geometry. Most of the learning-based methods require ground-truth labels~\cite{debortoli2019elevatenet,wang2021elevation,wang2022learning} which are difficult to obtain in oceanic scenarios. 

Very recently, self-supervised learning based on differentiable rendering and implicit neural representations has been introduced into 3D reconstruction from sonars, SSS~\cite{nils2022neural,xie2022sidescan} and FLS~\cite{qadri2023neural}. This framework frees the need for labeled training datasets and the use of implicit neural representations avoids the memory overhead associated with volumetric methods. However, Qadri~\textit{et al.}~\cite{qadri2023neural} focus on object reconstruction rather than seabed mapping which requires the vehicle hovering around the target to collect data. Although methods in~\cite{nils2022neural,xie2022sidescan} are designed for seabed mapping with standard surveys, they use gradient descent algorithms to find the corresponding elevation angle of the intersection between the arc and seafloor and subsequently a Lambertian model for intensity modeling. Such a surface-rendering approach can neither model the shadows nor the layover phenomenon~\cite{woock2010deep,xie2022towards}, where different elevation angles are projected to the same pixel in the sonar image. Besides, all of~\cite{nils2022neural,xie2022sidescan,qadri2023neural} use frequency encodings in implicit neural representations, which are non-parametric and thus neither adaptive nor efficient~\cite{muller2022instant}, limiting their ability for high-resolution bathymetry reconstruction. 

\begin{figure*}[!h]
\centering
\includegraphics[width=\linewidth]{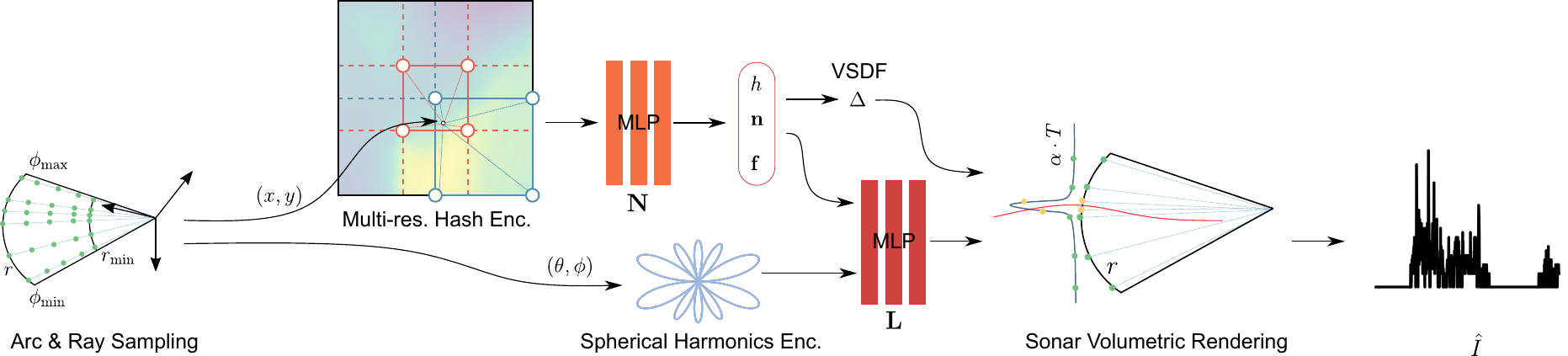}  
  \caption{\textbf{System Overview}: Given a range $r$ and azimuth angle $\theta$, we sample along the arc and for every sampled point on the arc, we sample along the acoustic ray. The spatial coordinates $(x,y)$ are encoded and then fed into the neural heightmap $\mathbf{N}$ outputting the height $h$, while at the same time the viewing angle $(\theta,\phi)$ is encoded using a spherical harmonics basis. The rendering network $\mathbf{L}$ takes spatial information, learned features $\mathbf{f}$, encoded viewing angles and surface normals $\mathbf{n}$ to learn the radiance field, which is later used together with the vertical signed distance $\Delta$ to apply sonar volumetric rendering to predict the returned intensity.}
  \label{fig:pipeline}
\end{figure*}

In this work, we address these shortcomings by proposing a differentiable rendering-based framework (Fig.~\ref{fig:pipeline}) that leverages the power of multi-resolution hash encodings~\cite{muller2022instant}, tailored to bathymetry reconstruction from FLS data during a standard survey with a "lawn-mower" pattern\footnote{A lawn-mower pattern that has the vehicle perform the survey as a series of long parallel lines.}. The contributions of this work are as follows: 
\begin{itemize}
\item We propose to use parametric multi-resolution hash encodings and a hierarchy sampling technique along the elevation arc that advance the efficiency of the volumetric renderer with low overhead, fast convergence and quick feedforward speed.
% \item We implement .
\item We model the beam pattern in both horizontal and vertical directions that can be jointly learned with bathymetry.
\item We evaluate the proposed method on both simulation and field data to illustrate the improvement over the state-of-the-art works. The implementation and simulation dataset are available in our Github\footnote{The source code link will be provided here upon acceptance.
}
\item We demonstrate how the proposed framework can combine FLS and MBES data for super-resolution mapping, leveraging the advantages of both
sensors.
\end{itemize}

\section{Related Works}

As aforementioned, different sparse 3D reconstructions~\cite{huang2016incremental,mai20173,westman2018feature,li2018pose,westman2019degeneracy} based on manually selected features~\cite{mai20173,huang2016incremental}, AKAZE features~\cite{westman2018feature,westman2019degeneracy} and learned features~\cite{li2018pose} have been introduced to FLS images and many of them focus more on simultaneous localization and mapping (SLAM)~\cite{huang2016incremental,westman2018feature,li2018pose,westman2019degeneracy} rather than on 3D reconstructions. 

As for dense 3D reconstructions, a number of works place restrictions or assumptions on the physical setup. Teixeira~\textit{et al.}~\cite{teixeira2016underwater,teixeira2019dense} use a FLS with $1^\circ$ elevation aperture to reconstruct a 3D model of a ship hull, prohibiting the method from extending to wide-aperture sonars. Westman~\textit{et al.}~\cite{westman2020theory} propose a non-light-of-sight (NLOS) method for 3D reconstruction on sonar based on Fermat paths, the use of which is considered to be an effective solution for the layover phenomenon. However, the method requires view rays perpendicular to the surface, which is impractical for bathymetry reconstruction. Wang~\textit{et al.}~\cite{wang2023motion} propose a method utilizing the motion field to estimate the missing elevation angle in a self-supervised learning framework, however, it places restrictions on the motion of the vehicle to avoid failures caused by  degenerate motions.

A different group of methods is based on shape-from-shading (SFS) techniques with generative models (Lambertian) assuming diffuse reflection. For object-based reconstruction, ~\cite{aykin20153,aykin2016modeling,westman2019wide} require estimates of object edges and/or shadows, making them unsuitable in real oceanic scenarios. For bathymetry reconstruction, neural FSF methods in~\cite{nils2022neural,xie2022sidescan} propose to use Sinusoidal Representation Networks (SIRENs)~\cite{siren2020} for the bathymetry representation and fit the sonar data from standard surveys through a gradient-based optimization. However, the use of the Lambertian model cannot explain the shadows in the sonar images thus introducing modeling errors especially during low-altitude surveys. As aforementioned, \cite{nils2022neural,xie2022sidescan} also assume there is only one elevation angle for each corresponding pixel in the sonar images, which ignores the layover phenomenon that happens commonly when the scene is complex such as surveying areas with boulders and rocks on the seafloor. Besides, the non-parametric frequency encoding used in SIRENs makes the representation inefficient, resulting in computational time and memory usage overhead.

Various supervised learning-based methods have also been proposed recently. DeBortoli~\textit{et al.}~\cite{debortoli2019elevatenet} propose to train a Convolutional Neural Network (CNN) on synthetic datasets and then fine-tune it on real datasets in a self-learning scheme, however the generalization ability and the sim-to-real gap limits such method's performance on field data. Xie~\textit{et al.}~\cite{xie2020inferring,xie2022bathymetric,xie2022neural} use MBES data to create ground truth to directly train a CNN on field data using supervised learning and show that the trained CNN can generalize to some extent with the same sensor setup. But the methods require collecting MBES to create a large training set every time the sensor setup is different. Similarly, Wang~\textit{et al.}~\cite{wang2021elevation,wang2022learning} have found that transferring acoustic view to pseudo front view helps the performance by considering the layover phenomenon, but such methods still require collecting real datasets with ground truth. 

Notably, inspired by NeuS~\cite{wang2021neus}, Qadri~\textit{et al.}~\cite{qadri2023neural} propose to replace the camera model with a sonar model and reconstruct a single object from multi-view FLS images, where a neural signed distance field (SDF) is used to represent the object. However, they rely on threshold-based filtering on the intensities to address the floater artifacts, similar to a background mask used in NeuS~\cite{wang2021neus}. Besides, they suffer from the same limitations as~\cite{nils2022neural} and~\cite{xie2022sidescan}, brought by frequency encodings in implicit neural representations. As for sampling along the arc, the lack of importance sampling makes it inefficient, especially when modeling wide-aperture sonars.

Implicit neural representations and differentiable rendering have gained much interest for 3D reconstruction, where NeRF~\cite{mildenhall2020nerf} made a breakthrough by using volumetric rendering to learn a neural radiance and density field, achieving state-of-the-art results.  Wang~\textit{et al.}~\cite{wang2021neus} propose to learn a neural SDF instead of the radiance and density field so that explicit surface constraints such as Eikonal loss can be added to achieve a high-quality surface reconstruction. Instead of using frequency encodings, which are non-parametric as in NeRF~\cite{mildenhall2020nerf} and NeuS~\cite{wang2021neus}, Instant-NGP~\cite{muller2022instant} proposes a multi-resolution hash encoding where a multi-resolution hash table of feature vectors is jointly trained with a smaller multi-layer perception (MLP), which increases the efficiency by reducing memory consumption and computational time.

In this work, leveraging the power of multi-resolution hash encodings, we represent the bathymetry with a neural heightmap, which suits better and is more efficient for the application of bathymetry reconstruction than volume-based shape representations. A volumetric differentiable renderer is used to render sonar intensities at a given pose, range and azimuth angle, where a hierarchy sampling scheme is used rather than stratified sampling to improve efficiency. With a kernel-based beam pattern modeling for both vertical and horizontal directions, the difference between rendered and measured intensities is used to optimize the bathymetry and beam patterns jointly in a self-supervised manner.

\section{Neural Volume Rendering for Reconstruction}
\subsection{FLS Model}
\begin{figure}[!h]
\centering
\includegraphics[width=\linewidth]{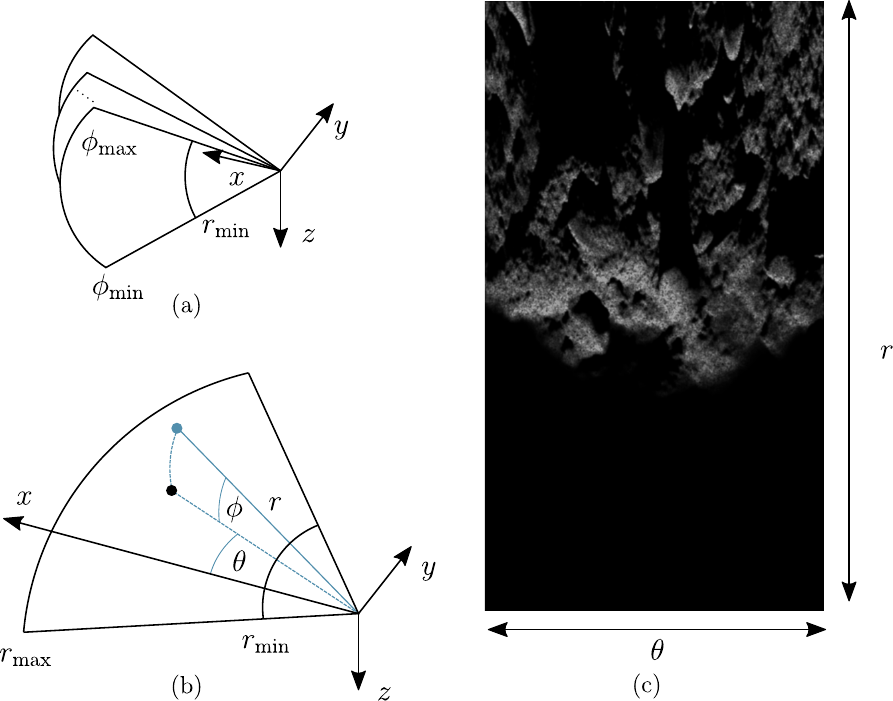}  
  \caption{(a) Each image column $\theta_i$ corresponds to a 2D fan-shape beam emitted from a transducer, recording returns between $r_\mathrm{min}$ and $r_\mathrm{max}$. (b) Geometry model of the FLS, where the blue point $(r,\theta,\phi)$ is a 3D point that is projected onto the image plane $z=0$. (c) An example of a sonar image in simulation. Each pixel at $(r,\theta)$ contains the returned intensities of all points along the elevation arc.}
  \label{fig:FLS_geometry}
\end{figure}
A multi-beam FLS's geometry can be described as in Fig.~\ref{fig:FLS_geometry}, where  each transducer in the azimuth direction $\mathit{\theta}$ emits a 2D fan-shaped beam and records the time-of-flight (TOF) of the sound waves and the returned back-scattered intensities, $\mathit{I}$. Each beam is very wide in the elevation direction $\mathit{\phi}$, where all the returns from the elevation arc ($\phi_{\mathrm{min}} \leq \phi \leq \phi_{\mathrm{max}}$) are projected onto the zero-elevation plane. This results in not only the ambiguity of the elevation angle for a given pixel corresponding with range $\mathit{r}$, azimuth angle $\mathit{\theta}$ with intensity $\mathit{I}$ but also the layover phenomenon where one bin consists of the returns from multiple 3D points along the elevation arc, if the geometry is complex. The returned intensity can be expressed as 
\begin{equation}\label{eq:intensity_modeling}
    I(r,\theta) = \int_{\phi_{\mathrm{min}}}^{\phi_{\mathrm{max}}} \beta(\theta,\phi)  \sigma(r,\theta,\phi)  T(r,\theta,\phi) L(r,\theta,\phi,\mathbf{v}) d\phi,
\end{equation}
where $T$ is the transmittance, $\sigma$ is the particle volume density, similarly as in \cite{wang2021neus}. $L$ is the radiance at a given 3D position $(r,\theta,\phi)$, from viewing direction $\mathbf{v}$, which depends on $\theta$, $\phi$ and sensor position. $\beta(\theta,\phi)$ models the beam pattern at given $\theta$ and $\phi$. 
\subsection{Efficient Neural Representations}
Similar to~\cite{wang2021neus} and~\cite{qadri2023neural}, we represent the surface and radiance field using two MLPs, except that for the surface, we leverage a neural heightmap $\mathbf{N}:\mathbb{R}^2\to\mathbb{R}$ that maps a 2D Cartesian position in world frame to its heights, $h=\mathbf{N}(x_w, y_w)$, instead of neural SDF to represent the bathymetry for the sake of efficiency. Instead of using non-parametric, frequency encoding as in~\cite{qadri2023neural}, we apply multi-resolution hash encoding~\cite{muller2022instant} to the spatial positions. Specifically, $L$ levels (two of which are shown as red and blue in Fig.~\ref{fig:pipeline}) of grids store trainable encoding parameters at the vertices of the grid. For a spatial position, the features at the corners are looked up from the corresponding hash tables and then linearly interpolated. The feature vectors for different levels are concatenated  before feeding into the MLP in $\mathbf{N}$.  For the neural radiance field $\mathbf{L}$, we encode the radiance associated with a point $\mathbf{x}\in\mathbb{R}^3$ and a viewing direction $\mathbf{v}\in\mathbb{S}^2$ given its vertical signed distance, surface normal and encoding feature vector. Note that, given a 3D point $(r,\theta,\phi)$ in polar coordinates in the local sonar frame, we need to transform it to Cartesian coordinates in the global reference frame $\mathbf{P}_p=[x_w, y_w, z_w]$ before calculating the vertical signed distance and normal. The transformation from polar coordinates to Cartesian coordinates is as follows:
\begin{align}
\begin{split}
    x_s &= r \cos(\theta) \cos(\phi)\\
    y_s &= r \sin(\theta) \cos(\phi)\\
    z_s &= r \sin(\phi),\\
\end{split}
\end{align}
and the transformation from local frame to world frame is
\begin{equation}
    [x_w, y_w, z_w]^T= \mathbf{R}^s_W  [x_s, y_s, z_s]^T + \mathbf{t}^s_W,
\end{equation}
given rotation matrix $\mathbf{R}^s_W$ and translation $\mathbf{t}^s_W$.

The vertical signed distance is simply $\Delta := z_w - \mathbf{N}(x_w,y_w)$ and the normal is calculated from the two gradient components $\nabla_x, \nabla_y$ of $\Delta$: 
\begin{equation}
    \mathbf{n}=[-\nabla_x\mathbf{N}(x_w,y_w),-\nabla_y\mathbf{N}(x_w,y_w), 1]^T.
\end{equation}

\subsection{Hierarchical Sampling}
\begin{figure}[!h]
\centering
\includegraphics[width=0.8\linewidth]{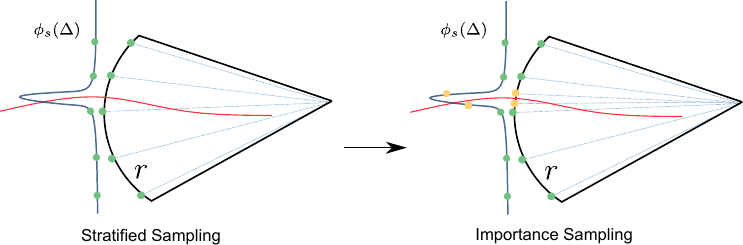}    \caption{Illustration of hierarchical sampling along the arc. Assuming the red curve is the seafloor, first $\mathbf{N_{\mathcal{A},s}}=5$ stratified samples (in green) are drawn on the arc and subsequently $\mathbf{N_{\mathcal{A},i}}=2$ importance samples (in yellow) are obtained using the S-density of the 5 green samples. The whole 7 samples are used to construct acoustic rays where we further sample along the rays to compute opacity and transmittance of the 7 samples along the arc.}
  \label{fig:Importance_sampling}
\end{figure}

As pointed out in~\cite{qadri2023neural}, to render sonar images, other than sampling points along the acoustic ray for corresponding pixels to compute $T$ and $\sigma$ in Eq.~\eqref{eq:intensity_modeling}, we also need to sample points along the elevation arc since we do not know where are the intersections between the arc and the surface. In this work, we use a hierarchical sampling strategy (illustrated in Fig.~\ref{fig:Importance_sampling}) when sampling along the arc by first stratified sampling $\mathbf{N_{\mathcal{A},s}}$ points along the elevation arc $\mathcal{A}_p$ and then conducting importance sampling to obtain another $\mathbf{N_{\mathcal{A},i}}$ elevation angles based on the coarse probability estimation, which is computed using the S-density $\phi_s(\Delta)=se^{-s\Delta}/(1+e^{-s\Delta})^2$ with standard deviations $s$. To compute the S-density, we need to feedforward these $\mathbf{N_{\mathcal{A},s}}$ samples to the neural heightmap $\mathbf{N}$, which increases computation overhead slightly. Note that, the S-density function is the derivative of the sigmoid function $\Phi_s(\Delta)=(1+e^{-s\Delta})^{-1}$. For the sampling along the ray, we follow a similar strategy as~\cite{qadri2023neural}, where the samples along the ray $\mathcal{R}_{p}$ contain exactly one point on the arc (at the end of the ray) and $\mathbf{N_{\mathcal{R}}}-1$ points uniformly sampled along the ray. 
\subsection{Beam Pattern}
To model the beam pattern, a characteristic property of the sensor, we follow the idea in~\cite{nils2022neural,xie2022sidescan} except that we estimate for both vertical and horizontal direction, $\beta(\theta,\phi)=\beta_\Theta(\theta)\beta_\Phi(\phi)$.  We use kernel densities whose kernel weights $\Theta_\mathcal{K},\Phi_\mathcal{K}$ are estimated jointly during the optimization.  We set kernels at fixed positions $\theta_{\mathcal{K}_1},\phi_{\mathcal{K}_1}$ evenly spread across the range:
\begin{align}
    \beta_{\Theta}(\theta)&=\sum_{\mathcal{K}_1}\Theta_{\mathcal{K}_1}\exp{(-\frac{(\theta_{\mathcal{K}_1}-\theta)^2}{2\sigma_{\Theta}^2})},\\
    \beta_{\Phi}(\phi)&=\sum_{\mathcal{K}_2}\Phi_{\mathcal{K}_2}\exp{(-\frac{(\phi_{\mathcal{K}_2}-\phi)^2}{2\sigma_{\Phi}^2})},
\end{align}
where $\sigma_{\Theta},\sigma_{\Phi}$ are the spreads of the kernels, set to be the range divided by the number of kernels $\mathcal{K}_1,\mathcal{K}_2$.
\subsection{Sonar Volumetric Rendering}
The discrete form of intensity modeling in Eq.~\eqref{eq:intensity_modeling} is:
\begin{equation}
    \hat{I}(r,\theta)=\sum_{\mathbf{P}_p\in\mathcal{A}_p}\beta(\mathbf{P}_p)T(\mathbf{P}_p)\alpha(\mathbf{P}_p)\mathbf{L}(\mathbf{P}_p),
\end{equation}
where $\mathcal{A}_p$ is the elevation arc located at $(r,\theta)$, $\mathbf{L}(\mathbf{P}_p)$ is the predicted intensity at $\mathbf{P}_p$ predicted by the neural renderer $\mathbf{L}$, $\alpha(\mathbf{P}_p)$ is the discrete opacity at $\mathbf{P}_p$ which can be computed following~\cite{wang2021neus}:
\begin{equation}
    \alpha(\mathbf{p}_i)=\mathrm{max}(\frac{\mathbf{\Phi}_s(\Delta(\mathbf{p}_i))-\mathbf{\Phi}_s(\Delta(\mathbf{p}_{i+1}))}{\mathbf{\Phi}_s(\Delta(\mathbf{p}_i))},0),
\end{equation}
where $\mathbf{p}_{i},\mathbf{p}_{i+1}$ are consecutive samples along the acoustic ray. Finally we have the discrete transmittance at the endpoint of the ray $\mathbf{P}_p$ shown to equal~\cite{qadri2023neural}:
\begin{equation}
T(\mathbf{P}_p)=\prod_{\mathbf{p}\in\mathcal{R}_{\mathbf{P}_p}\smallsetminus\mathbf{P}_p}(1-\alpha(\mathbf{p})),
\end{equation}
where $\mathcal{R}_{\mathbf{P}_p}$ is the acoustic ray ends at $\mathbf{P}_p$ and the transmittance is computed using the set of points on that ray excluding the endpoint.

\subsection{Loss Functions}
The loss function consists of two terms, the intensity loss
\begin{equation}
    \mathcal{L}_{\mathrm{int}}=\frac{1}{|\mathcal{P}|}\sum_{p\in\mathcal{P}} \left \lVert \hat{I}(p)-I(p) \right \rVert_1
\end{equation}

and the regularization term
\begin{equation}
     \mathcal{L}_{\mathrm{reg}}=\frac{1}{|\mathcal{P}|}\sum_{p\in\mathcal{P}} (\left \lVert \mathbf{n}(p) \right \rVert_2 -1)^2
\end{equation}
to encourage surface smoothness. $\mathcal{P}$ is the set of sampled pixels in a mini-batch. Optionally an altimeter loss could be added to accelerate convergence:
\begin{equation}
    \mathcal{L}_{\mathrm{alt}}=\frac{1}{|\mathcal{P}_{\mathrm{alt}}|}\sum_{p\in\mathcal{P}_{\mathrm{alt}}} \left \lVert\Delta(p) \right \rVert_1,
\end{equation}
where $\mathcal{P}_{\mathrm{alt}}$ is the set of single-beam altimeter readings along the ROV transit. The total loss is a weighted sum of the loss terms above.
\section{Experiments}

\begin{table}[t]
\caption{Dataset Details}
\label{tab:dataset}
\begin{tabular}{llll}
\hline
Dataset                & Aerial Rocks & Sponge Ridge 1 & Sponge Ridge 2 \\ \hline
Vehicle                & Simulator     & Ventana+LASS      & Ventana              \\
Avg altitude (m)       & 4.84          & 3.67         &      3.78         \\
Image res.               & 919$\times$512    & 1887$\times$512   &    920$\times$512    \\
Survey area            & $\sim$70m$\times$70m     & $\sim$80m$\times$80m      & $\sim$80m$\times$40m            \\
No. of images            & 42354         &    12851      &     9302         \\
FLS range (m)     & 30            & 31           & 30                \\
Total traj. (m)         & 408.75        &   387.63     &      365.05       \\ 
Duration (min) & 71 (10Hz)     &  38 (5Hz)   &  10 (15Hz)        
\end{tabular}
\end{table}
In this section, we evaluate the performance of the proposed method on both simulation and field datasets from low-altitude surveys. We demonstrate two applications, namely bathymetry reconstruction from FLS and super-resolution bathymetric mapping from FLS with the aid of a low-resolution prior map. For the former application, we use the simulation dataset and one of the field datasets, \textcolor{black}{Sponge Ridge 1}. We compare the proposed model against three baselines:  1) \textit{Lambert.$+$GD}: a Lambertian model is used for the intensity modeling and gradient descent (GD) to compute the one intersection between the elevation arc and the seabed as in~\cite{nils2022neural,xie2022sidescan}. A threshold set empirically is used to remove shadows in the loss computation. 2) \textit{Lambert.$+$Sampl.}: the same Lambertian model for the sonar scattering modeling but only using stratify sampling strategy as in~\cite{qadri2023neural} is used. 3) \textit{Freq.$+$Sampl.}: the full model in~\cite{qadri2023neural} with the exception that we use a neural heightmap instead of a full 3D SDF. For the super-resolution mapping application, we use the other field dataset, \textcolor{black}{Spondge Ridge 2} where we have a prior bathymetric map from an AUV survey at a much higher altitude. We use the gridded bathymetric data from the low-resolution map (1 m) in the altimeter loss to directly supervise the neural heightmap and demonstrate that with the aid of FLS from low-altitude surveys, we can create a super-resolution (5 cm) bathymetry with high fidelity.

\subsection{Data}
\subsubsection{Simulation Data: Aerial Rocks}
The simulation dataset was generated using the Stonefish~\cite{cieslak2019stonefish} where the seabed terrain is based on public heightmaps to simulate a rocky area. The ROV was programmed to survey in a variation of the common lawn-mower trajectory~\cite{munoz2024learning} (seen Fig.~\ref{fig:Sim_bathy}) with specifications shown in Table~\ref{tab:dataset}.

 \subsubsection{Field Data} Two field datasets were collected using the ROV Ventana aboard the R/V Rachel Carson from the Monterey Bay Aquarium Research Institute (MBARI), located in Moss Landing, California. \textcolor{black}{Sponge Ridge 1} was collected with the Low Altitude Survey System (LASS) mounted on ROV Ventana during an expedition in 2023 in Monterey Bay. The navigation was controlled by a scripted mission plan using a 10 m line spacing. A Kearfott SeaDevil inertial navigation system (INS) aided by a RDI Workhorse Doppler Velocity Log (DVL) is used for navigation estimation. Data from the Reson SeaBat 7125 MBES on LASS is used to provide 10 cm resolution map as ground truth.  
 \textcolor{black}{Sponge Ridge 2} was collected with only ROV Ventana during an expedition in 2022 in Monterey Bay, where a pilot was flying the ROV. An Octans Fiber Optic Gyrode and a RDI Workhouse DVL combo is used to estimate the 6 degree-of-freedom (DOF) pose of the ROV, which is treated as the ground truth for navigation. A prior map with 1 m resolution collected with MBARI mapping AUV with MBES is available as a reference. 

 For both datasets, a GPS receiver is used to initialize the navigation prior to deployment and ultra-short baseline (USBL) tracking is used to stabilize the navigation fix during ROV descent. Both datasets were collected using a Gemini 720is FLS with a slightly different setup, as shown in Table~\ref{tab:dataset}. 
Both simulation and field sonar have a horizontal field of view (HFOV) of 120$^\circ$ and vertical field of view (VFOV) of 20$^\circ$. For both simulation and field data, the ROV were running low-altitude surveys, as shown in Table~\ref{tab:dataset}.

\subsection{Metrics}
To evaluate the performance, we use the mean absolute errors (MAE) and the standard deviation (STD) of the signed errors between the reconstructed bathymetry and ground truth, both in meters. Additionally, we also computed the structural similarity index measure (SSIM) between the reconstructed bathymetry and ground truth by treating the heightmaps as 16-bit gray-scale images. SSIM, $[0,1]$, is a metric for evaluating the quality of the reconstructed images, with 1 indicating the same as the reference image.

\subsection{Implementation Details}
For simulation and field datasets, we use an MLP with 2 hidden layers of size 64 with multi-resolution hash encoding to model $\mathbf{N}$. The hash table size is $2^{15}$ ($L=15$) and the maximum resolution $N_{\mathrm{max}}=1024$. For $\mathbf{L}$, we again use a 2-hidden-layer MLP of 64 neurons wide but with sphere harmonics encoding (up to 3 degrees) for the view directions. For the baselines that use frequency encodings, we use 6 frequencies and keep the same sphere harmonics encoding to
view directions with up to 3 degrees. For simulation experiments, we did not estimate the beam pattern but for experiments on field datasets, we use $\mathcal{K}_1=30$ kernels over 120$^\circ$ for the horizontal beam pattern modeling and $\mathcal{K}_2=10$ kernels over 40$^\circ$ for vertical beam pattern, given the 3dB beam width of Gemini 720is is 20$^\circ$. As for the hierarchical sampling, for simulation dataset, we assume 20$^\circ$ vertical sensor opening and we first sample $\mathbf{N_{\mathcal{A},s}}=15$ points along the elevation arc and then another $\mathbf{N_{\mathcal{A},i}}=15$ importance samples. For both field datasets, we assume 40$^\circ$ sensor opening with $\mathbf{N_{\mathcal{A},s}}=30$ and  $\mathbf{N_{\mathcal{A},i}}=30$. We set $\mathbf{N_{\mathcal{R}}}=60$ for samples along each ray in all experiments. All models are trained for 60 k steps with Adam optimizer with a learning rate $5\times10^{-2}$ that linearly decays by a factor of 0.97 every 600 steps. For each mini-batch we randomly select one beam out of 512. To enforce and accelerate convergence, we conduct Progressive Training strategy as in~\cite{wang2023neus2} and we also use the altimeter loss. The altimeter readings are simulated for the simulation dataset and as for the \textcolor{black}{Sponge Ridge 1} dataset, we use the actual readings from the onboard DVL. For the super-resolution mapping experiment, we convert the gridded low-resolution prior map to point clouds, simulating dense altimeter readings.
\begin{figure}[h!]
\centering
\includegraphics[width=\linewidth]{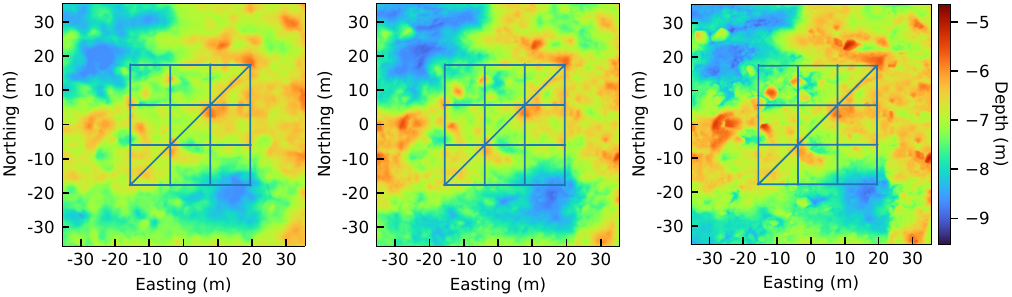}  
  \caption{Reconstructed bathymetry for \textit{Freq.$+$Sampl.} (left) and for the proposed method (middle) and ground truth (right) for simulation dataset, at 10 cm resolution (ROV trajectory in blue).}
  \label{fig:Sim_bathy}
\end{figure}
\section{Results}
\subsection{Bathymetry Reconstruction} 
\subsubsection{Simulation dataset results}

\begin{table}[]
\caption{Simulation Quantitative Results}
\label{tab:simulation}
\resizebox{\columnwidth}{!}{\begin{tabular}{l|ccc|ccc}
\hline
\multirow{2}{*}{\backslashbox{Model}{Area}} & \multicolumn{3}{c|}{70$\times$70 {[}m{]}} & \multicolumn{3}{c}{50$\times$50 {[}m{]}} \\\cline{2-7}
                  & MAE $\downarrow$    & STD $\downarrow$   & SSIM $\uparrow$  & MAE $\downarrow$    & STD $\downarrow$   & SSIM $\uparrow$   \\\hline
Lambert.+GD       & 0.305   & 0.209  & 0.875  & 0.254   & 0.209  & 0.863  \\\hline
Lambert.+Sampl.   & 0.385   & 0.355  & 0.799  & 0.241   & 0.212  & 0.812  \\\hline
Freq.+Sampl. & 0.282   & 0.235  & 0.813  & 0.210   & 0.200  & 0.818  \\\hline
Ours              & \textbf{0.240}   & \textbf{0.187}  & \textbf{0.905}  & \textbf{0.133}   & \textbf{0.115}  & \textbf{0.906} \\\hline
\end{tabular}}
\end{table}
The MAE, STD and SSIM for the simulation datasets computed for all aforementioned models are shown in Table~\ref{tab:simulation}. In addition to the metrics in the survey area (70$\times$70 m) we also compute the metrics in the inner area (50$\times$50 m). Of all the models, the proposed method has the best performance in all categories, demonstrating the importance of relaxation on the Lambertian assumption as well as the advantages of parametric encodings that could automatically focus on relevant details with similar amount of trainable parameters (MLP weights + encoding parameters). Note that across all models, the errors in the inner area are lower than the whole survey area, which are congruent with our expectations since at the outer area, the number of observations from the sonar is lower, resulting in weaker constraints. Furthermore, around the perimeter, the increased distance between the seabed and the sonar, results in higher noise levels and lower resolution, inevitably leading to lower quality reconstruction. This creates a trade-off between efficiency (survey coverage) and reconstruction quality. The predicted bathymetry at 10 cm resolution from baseline \textit{Freq.+Sampl.} (left) and our method (middle) together with the ground truth (right) are shown in Fig.~\ref{fig:Sim_bathy}. The plot shows that the baseline method manages to reconstruct the topographic details of the surveyed rocky area fairly well but our approach is able to capture finer geometric details with higher fidelity, providing a sharper heightmap. This is consistent with the SSIM metric where ours (0.905) has a 11.3\% improvement compared to the baseline (0.813). Furthermore, we can observe from Fig.~\ref{fig:Sim_bathy} that the outer area, for example at the upper left corner and lower right corner,  the baseline \textit{Freq.+Sampl.} looses much more details than the proposed method, mainly because of the limited non-parametric frequency encodings.

\subsubsection{\textcolor{black}{Sponge Ridge 1}}
\begin{figure}[!h]
\centering
\includegraphics[width=\linewidth]{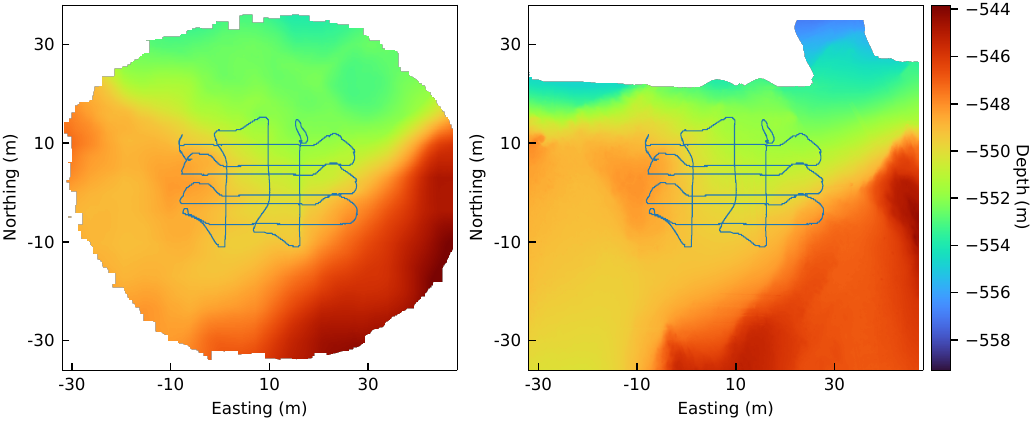}  
  \caption{Reconstructed bathymetry for the proposed method (left) and ground truth (right) for \textcolor{black}{Sponge Ridge 1} dataset collected with LASS, at 10 cm resolution (ROV trajectory in blue).}
  \label{fig:LASS_bathy}
\end{figure}

\begin{table}[]
\caption{Field Results - Absolute Error }
\label{tab:field}
\begin{tabular}{l|cc|cc}
\hline
\multirow{2}{*}{\backslashbox{Model}{Area}} & \multicolumn{2}{c|}{80$\times$80 {[}m{]}} & \multicolumn{2}{c}{60$\times$60 {[}m{]}} \\ \cline{2-5}
                  & MAE   $\downarrow$     & STD  $\downarrow$       & MAE $\downarrow$       & STD  $\downarrow$       \\\hline
Lambert.+GD       & 0.922       & 0.838       & 0.808       & 0.785       \\\hline
Lambert.+Sampl.   & 0.810       & 0.593       & 0.738       & 0.570       \\\hline
Freq.+Sampl.      & 0.847       & 0.635       & 0.714       & 0.582       \\\hline
Ours$^{-}$      & 0.727       & 0.604       & 0.643       & 0.515       \\\hline
Ours              & \textbf{0.531}       & \textbf{0.495}       & \textbf{0.450}       & \textbf{0.464}      \\\hline
\end{tabular}
\end{table}
Table~\ref{tab:field} shows the MAE and STD for the \textcolor{black}{Sponge Ridge 1} dataset. All models except \textit{Ours$^-$} use the altimeter data from the onboard DVL for bathymetric constraints and we estimate the beam pattern the same way across all models. For the setup in \textit{Ours$^-$}, we only used intensity and regularization loss. SSIM is not computed since we do not have full-coverage ground truth bathymetry, as shown in Fig.~\ref{fig:LASS_bathy} (right). Again the best results are obtained from the proposed method, with 0.531 m MAE for the whole survey area and 0.450 m at the inner area, demonstrating the proposed method's advantages over the baselines. Furthermore, compared to Lambertian-based methods, our approach relaxes the pure diffusion reflectance assumption and is able to better model the sonar ensonification process. Note that with the proposed method without altimeter loss, i.e., {Ours$^-$}, the MAE is still lower than all baselines. Fig.~\ref{fig:LASS_bathy} shows the reconstructed bathymetry from the proposed approach gridded at 10 cm resolution, showing that the shape of the reconstructed ridges. As one can notice that the reconstructed quality of the topology deteriorates as it gets further away from sonar. This is more noticeable than the case in simulation due to the higher level noise in field data. Note that we masked out the places that have been ensonified less than twice, which are generally at the perimeter. The plot and the table demonstrate the quality of the bathymetry we could obtain from field FLS using the proposed approach.  

\begin{figure}[h]
\centering
\includegraphics[width=\linewidth]{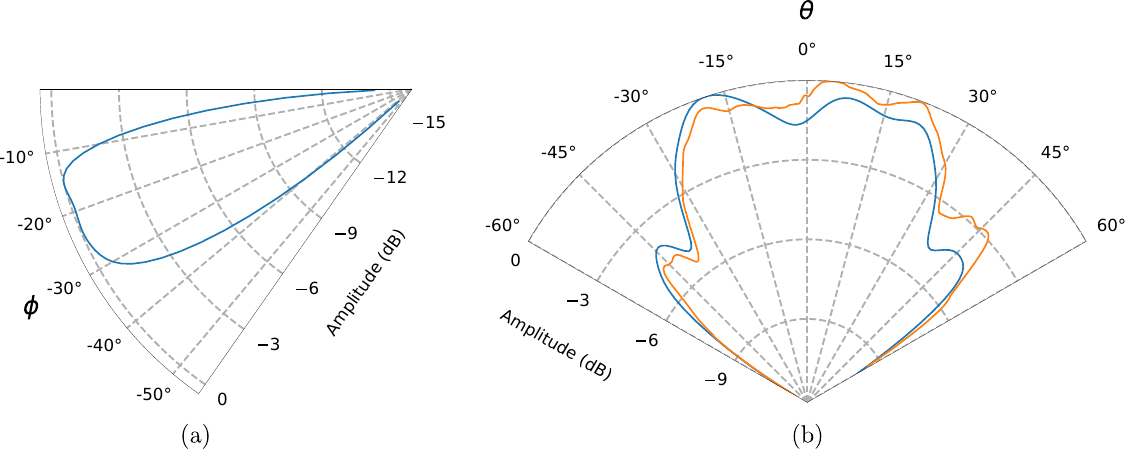}  
  \caption{Normalized beam pattern. (a) Estimated beam pattern in vertical direction. (b) Beam pattern in horizontal direction, where the blue line is estimated during the optimization and the orange line is the average intensities over azimuth angles across all FLS images in the dataset, as a crude reference.}
  \label{fig:LASS_bP}
\end{figure}

Fig.~\ref{fig:LASS_bP} shows the beam pattern estimated together with the bathymetry during the optimization. The estimated vertical beam pattern [Fig.~\ref{fig:LASS_bP}(a)] is modelled from $\phi_{\mathrm{max}}=-5^\circ$ to $\phi_{\mathrm{min}}=-45^\circ$, where from the plot we can see the 3dB beam width is roughly 20$^\circ$, which agrees with the datasheet of Gemini 720is. Fig.~\ref{fig:LASS_bP}(b) shows the estimated horizontal beam pattern and the ``calibration'', which is obtained by averaging all intensities over azimuth angles for the entire dataset, excluding the water column parts. This is a crude approximation of the calibration, which can be used to qualitatively evaluate the estimated beam pattern. 

\subsection{Super Resolution Mapping}
\begin{figure}[!h]
\centering
\includegraphics[width=\linewidth]{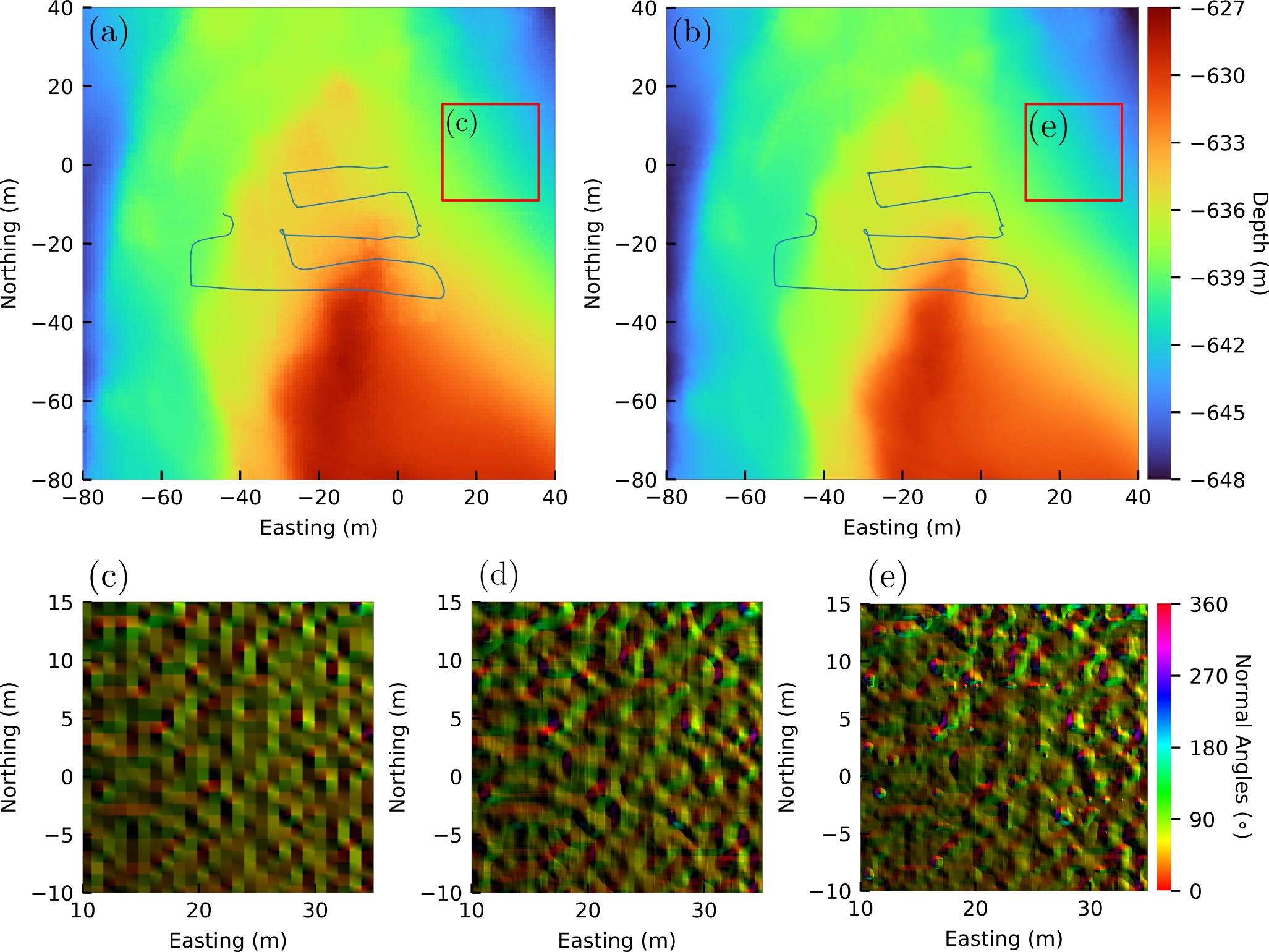}  
  \caption{Bathymetry (top) for the low-resolution prior map (a) and the reconstructed super-resolution map using proposed method gridded at 5 cm resolution (b) for \textcolor{black}{Sponge Ridge 2} dataset collected with Ventana (ROV trajectory in blue). At the bottom we zoom in and show the gradients of the map (the red square) for interpolation of prior map in (c),  training NNs only with point clouds from the prior map in (d), and training NNs with FLS and point clouds from the prior map in (e). }
  \label{fig:Ventana_sr_bathy_merged}
\end{figure}

Fig.~\ref{fig:Ventana_sr_bathy_merged} shows the estimated bathymetry gridded at 5 cm resolution (b) and the prior map with 1 m resolution (a). It is difficult to see the effect of super resolution from heightmaps directly, other than the prior map is more pixelated compared to our estimation. To demonstrate the details of the super resolution bathymetry, we also compute the gradients of the heightmaps [Fig.~\ref{fig:Ventana_sr_bathy_merged} (c)-(e)]. Note that since the estimated bathymetry is represented by an MLP, we can compute the exact gradients. As for the gradient map for the prior map, we use finite difference to approximate the gradients after interpolating the map to 5 cm resolution. The gradient maps reveal that the reconstructed bathymetry not only maintains the shape of the general structure but also shows many details such as small rocks and ripples on the seafloor. The plot reflects that the high resolution FLS images due to its high frequency and low altitudes provide rich information on the micro-topology of the seabed which is not available from MBES data from AUVs flying at higher altitudes.  Note that Fig.~\ref{fig:Ventana_sr_bathy_merged} (d) illustrates that if we directly fit the prior map to the MLP, we could leverage the continuous property of implicit neural representations and obtain a heightmap with sharper details [(d)] than interpolating the gridded data [(c)]. However, rocks that are smaller than MBES's footprint can only be resolved with the aid of FLS data, as shown in (e).

\section{Conclusions and Future Work}
We have presented a learning-based approach based on neural rendering for reconstructing bathymetry from FLS data in a self-supervising manner. The proposed framework has been tested on three surveys, two of them collected with field robots. When compared to the current state-of-the-art solutions for bathymetry reconstruction from imaging sonars, our method is capable of estimating high-resolution bathymetry with lower errors and higher fidelity, thanks to the novel contributions proposed in our approach. The results have shown that the proposed method could be applied to FLS field data from standard surveys for bathymetric mapping. Additionally the super-resolution experiment shows the application of using FLS from ROVs or small AUVs in low-altitude surveys to increase the resolution of bathymetry from a higher altitude survey, from for example, large AUVs or surface vessels.

The current major limitation of the proposed method is that we assume to have very accurate navigation estimates, which rarely holds in long missions if only relying on dead reckoning (DR). The unbounded drift over time will limit the quality of the reconstructing map. More research focusing on FLS SLAM could help to reduce the drift in the DR estimates, which could be used prior to applying the proposed method for seabed mapping. Another possible future work is to incorporate the neural rendering-based mapping into the FLS SLAM framework, for example, commonly a tracking thread to estimate the $SE(3)$ pose of the sonar and a parallel mapping thread to build the bathymetric maps only using key frames. Another possible future work is that, if a prior map is available, how to use FLS for localization and substantially creating a super-resolution bathymetry. 

Another limitation is that our method is mostly suited to offline optimization. One of the key reasons is that we need to sample along the elevation arc, resulting in more samples being fed into the MLPs and subsequently their gradient computation. This increases the training time compared to works using camera images for 3D reconstruction in real time~\cite{muller2022instant}. However, recent advances based on rasterization rather than ray casting such as 3D Gaussian Splatting~\cite{kerbl20233d} are significantly faster than methods based on~\cite{muller2022instant}, and can potentially be used for FLS mapping or even SLAM~\cite{Matsuki:Murai:etal:CVPR2024} in real time.

\section*{ACKNOWLEDGMENT}
The authors gratefully thank Bastián Muñoz for creating the simulation data. The authors also acknowledge the Monterey Bay Aquarium Research Institute (MBARI) engineering and technical team, 
especially Dave Caress, Kent Headley, Eric Martin, Kevin Barnard and Sebastián Rodriguez, 
the R/V Rachel Carson officers and crew for their support in collecting the field data. This work was a contribution of the CoMPAS laboratory at MBARI and was supported by the David and Lucile Packard Foundation.  The network training was partly enabled by the Berzelius resource provided by the Knut and Alice Wallenberg Foundation at the National Supercomputer Centre.

\bibliographystyle{IEEEtran}
\bibliography{IEEEabrv,root}

\begin{thebibliography}{10}
\providecommand{\url}[1]{#1}
\csname url@samestyle\endcsname
\providecommand{\newblock}{\relax}
\providecommand{\bibinfo}[2]{#2}
\providecommand{\BIBentrySTDinterwordspacing}{\spaceskip=0pt\relax}
\providecommand{\BIBentryALTinterwordstretchfactor}{4}
\providecommand{\BIBentryALTinterwordspacing}{\spaceskip=\fontdimen2\font plus
\BIBentryALTinterwordstretchfactor\fontdimen3\font minus \fontdimen4\font\relax}
\providecommand{\BIBforeignlanguage}[2]{{%
\expandafter\ifx\csname l@#1\endcsname\relax
\typeout{** WARNING: IEEEtran.bst: No hyphenation pattern has been}%
\typeout{** loaded for the language `#1'. Using the pattern for}%
\typeout{** the default language instead.}%
\else
\language=\csname l@#1\endcsname
\fi
#2}}
\providecommand{\BIBdecl}{\relax}
\BIBdecl

\bibitem{huang2016incremental}
T.~A. Huang and M.~Kaess, ``Incremental data association for acoustic structure from motion,'' in \emph{Proc. IEEE/RSJ Int. Conf. Intell. Robots Syst.}\hskip 1em plus 0.5em minus 0.4em\relax IEEE, 2016, pp. 1334--1341.

\bibitem{mai20173}
N.~T. Mai, H.~Woo, Y.~Ji, Y.~Tamura, A.~Yamashita, and H.~Asama, ``{3-D} reconstruction of underwater object based on extended {Kalman} filter by using acoustic camera images,'' \emph{IFAC-PapersOnLine}, vol.~50, no.~1, pp. 1043--1049, 2017.

\bibitem{westman2018feature}
E.~Westman, A.~Hinduja, and M.~Kaess, ``Feature-based {SLAM} for imaging sonar with under-constrained landmarks,'' in \emph{Proc. IEEE Int. Conf. Robot. Autom.}\hskip 1em plus 0.5em minus 0.4em\relax IEEE, 2018, pp. 3629--3636.

\bibitem{li2018pose}
J.~Li, M.~Kaess, R.~M. Eustice, and M.~Johnson-Roberson, ``Pose-graph {SLAM} using forward-looking sonar,'' \emph{IEEE Robot. Automat. Lett.}, vol.~3, no.~3, pp. 2330--2337, 2018.

\bibitem{westman2019degeneracy}
E.~Westman and M.~Kaess, ``Degeneracy-aware imaging sonar simultaneous localization and mapping,'' \emph{{IEEE} J. Ocean. Eng.}, vol.~45, no.~4, pp. 1280--1294, 2019.

\bibitem{teixeira2016underwater}
P.~V. Teixeira, M.~Kaess, F.~S. Hover, and J.~J. Leonard, ``Underwater inspection using sonar-based volumetric submaps,'' in \emph{Proc. IEEE/RSJ Int. Conf. Intell. Robots Syst.}\hskip 1em plus 0.5em minus 0.4em\relax IEEE, 2016, pp. 4288--4295.

\bibitem{teixeira2019dense}
P.~V. Teixeira, D.~Fourie, M.~Kaess, and J.~J. Leonard, ``Dense, sonar-based reconstruction of underwater scenes,'' in \emph{Proc. IEEE/RSJ Int. Conf. Intell. Robots Syst.}\hskip 1em plus 0.5em minus 0.4em\relax IEEE, 2019, pp. 8060--8066.

\bibitem{westman2020theory}
E.~Westman, I.~Gkioulekas, and M.~Kaess, ``A theory of fermat paths for {3D} imaging sonar reconstruction,'' in \emph{Proc. IEEE/RSJ Int. Conf. Intell. Robots Syst.}\hskip 1em plus 0.5em minus 0.4em\relax IEEE, 2020, pp. 5082--5088.

\bibitem{wang2023motion}
Y.~Wang, Y.~Ji, C.~Wu, H.~Tsuchiya, H.~Asama, and A.~Yamashita, ``Motion degeneracy in self-supervised learning of elevation angle estimation for {2D} forward-looking sonar,'' in \emph{Proc. IEEE/RSJ Int. Conf. Intell. Robots Syst.}\hskip 1em plus 0.5em minus 0.4em\relax IEEE, 2023, pp. 6133--6140.

\bibitem{aykin20153}
M.~D. Aykin and S.~Negahdaripour, ``On {3-D} target reconstruction from multiple {2-D} forward-scan sonar views,'' in \emph{Proc. IEEE OCEANS Conf.}\hskip 1em plus 0.5em minus 0.4em\relax IEEE, 2015, pp. 1--10.

\bibitem{aykin2016modeling}
M.~D. Aykin and S.~S. Negahdaripour, ``Modeling {2-D} lens-based forward-scan sonar imagery for targets with diffuse reflectance,'' \emph{{IEEE} J. Ocean. Eng.}, vol.~41, no.~3, pp. 569--582, 2016.

\bibitem{westman2019wide}
E.~Westman and M.~Kaess, ``Wide aperture imaging sonar reconstruction using generative models,'' in \emph{Proc. IEEE/RSJ Int. Conf. Intell. Robots Syst.}\hskip 1em plus 0.5em minus 0.4em\relax IEEE, 2019, pp. 8067--8074.

\bibitem{guerneve2015underwater}
T.~Guerneve and Y.~Petillot, ``Underwater {3D} reconstruction using {BlueView} imaging sonar,'' in \emph{Proc. IEEE OCEANS Conf.}\hskip 1em plus 0.5em minus 0.4em\relax IEEE, 2015, pp. 1--7.

\bibitem{guerneve2018three}
T.~Guerneve, K.~Subr, and Y.~Petillot, ``Three-dimensional reconstruction of underwater objects using wide-aperture imaging sonar,'' \emph{J. Field Robot.}, vol.~35, no.~6, pp. 890--905, 2018.

\bibitem{wang20183d}
Y.~Wang, Y.~Ji, H.~Woo, Y.~Tamura, A.~Yamashita, and A.~Hajime, ``{3D} occupancy mapping framework based on acoustic camera in underwater environment,'' \emph{IFAC-PapersOnLine}, vol.~51, no.~22, pp. 324--330, 2018.

\bibitem{westman2020volumetric}
E.~Westman, I.~Gkioulekas, and M.~Kaess, ``A volumetric albedo framework for {3D} imaging sonar reconstruction,'' in \emph{Proc. IEEE Int. Conf. Robot. Autom.}\hskip 1em plus 0.5em minus 0.4em\relax IEEE, 2020, pp. 9645--9651.

\bibitem{debortoli2019elevatenet}
R.~DeBortoli, F.~Li, and G.~A. Hollinger, ``Elevatenet: A convolutional neural network for estimating the missing dimension in {2D} underwater sonar images,'' in \emph{Proc. IEEE/RSJ Int. Conf. Intell. Robots Syst.}\hskip 1em plus 0.5em minus 0.4em\relax IEEE, 2019, pp. 8040--8047.

\bibitem{wang2021elevation}
Y.~Wang, Y.~Ji, D.~Liu, H.~Tsuchiya, A.~Yamashita, and H.~Asama, ``Elevation angle estimation in {2D} acoustic images using pseudo front view,'' \emph{IEEE Robot. Automat. Lett.}, vol.~6, no.~2, pp. 1535--1542, 2021.

\bibitem{wang2022learning}
Y.~Wang, Y.~Ji, H.~Tsuchiya, H.~Asama, and A.~Yamashita, ``Learning pseudo front depth for {2D} forward-looking sonar-based multi-view stereo,'' in \emph{Proc. IEEE/RSJ Int. Conf. Intell. Robots Syst.}\hskip 1em plus 0.5em minus 0.4em\relax IEEE, 2022, pp. 8730--8737.

\bibitem{nils2022neural}
N.~Bore and J.~Folkesson, ``Neural shape-from-shading for survey-scale self-consistent bathymetry from sidescan,'' \emph{{IEEE} J. Ocean. Eng.}, vol.~48, no.~2, pp. 416--430, 2023.

\bibitem{xie2022sidescan}
Y.~Xie, N.~Bore, and J.~Folkesson, ``Sidescan only neural bathymetry from large-scale survey,'' \emph{Sensors}, vol.~22, no.~14, p. 5092, 2022.

\bibitem{qadri2023neural}
M.~Qadri, M.~Kaess, and I.~Gkioulekas, ``Neural implicit surface reconstruction using imaging sonar,'' in \emph{Proc. IEEE Int. Conf. Robot. Autom.}\hskip 1em plus 0.5em minus 0.4em\relax IEEE, 2023, pp. 1040--1047.

\bibitem{woock2010deep}
P.~Woock and C.~Frey, ``Deep-sea {AUV} navigation using side-scan sonar images and {SLAM},'' in \emph{Proc. IEEE OCEANS Conf.}\hskip 1em plus 0.5em minus 0.4em\relax IEEE, 2010, pp. 1--8.

\bibitem{xie2022towards}
Y.~Xie, N.~Bore, and J.~Folkesson, ``Towards differentiable rendering for sidescan sonar imagery,'' in \emph{Proc. IEEE/OES Auton. Underwater Veh. Symp.}\hskip 1em plus 0.5em minus 0.4em\relax IEEE, 2022, pp. 1--6.

\bibitem{muller2022instant}
T.~M{\"u}ller, A.~Evans, C.~Schied, and A.~Keller, ``Instant neural graphics primitives with a multiresolution hash encoding,'' \emph{ACM Trans. Graph.}, vol.~41, no.~4, pp. 1--15, 2022.

\bibitem{siren2020}
V.~Sitzmann, J.~Martel, A.~Bergman, D.~Lindell, and G.~Wetzstein, ``Implicit neural representations with periodic activation functions,'' \emph{Adv. Neural Inf. Process. Syst.}, vol.~33, 2020.

\bibitem{xie2020inferring}
Y.~Xie, N.~Bore, and J.~Folkesson, ``Inferring depth contours from sidescan sonar using convolutional neural nets,'' \emph{IET Radar, Sonar \& Navigation}, vol.~14, no.~2, pp. 328--334, 2020.

\bibitem{xie2022bathymetric}
------, ``Bathymetric reconstruction from sidescan sonar with deep neural networks,'' \emph{{IEEE} J. Ocean. Eng.}, vol.~48, no.~2, pp. 372--383, 2022.

\bibitem{xie2022neural}
------, ``Neural network normal estimation and bathymetry reconstruction from sidescan sonar,'' \emph{{IEEE} J. Ocean. Eng.}, vol.~48, no.~1, pp. 218--232, 2022.

\bibitem{wang2021neus}
P.~Wang, L.~Liu, Y.~Liu, C.~Theobalt, T.~Komura, and W.~Wang, ``Neus: Learning neural implicit surfaces by volume rendering for multi-view reconstruction,'' \emph{Adv. Neural Inf. Process. Syst.}, vol.~34, pp. 27\,171--27\,183, 2021.

\bibitem{mildenhall2020nerf}
B.~Mildenhall, P.~P. Srinivasan, M.~Tancik, J.~T. Barron, R.~Ramamoorthi, and R.~Ng, ``{NeRF}: Representing scenes as neural radiance fields for view synthesis,'' in \emph{Proc. Eur. Conf. Comput. Vis.}, 2020, pp. 405--421.

\bibitem{cieslak2019stonefish}
P.~Cie{\'s}lak, ``Stonefish: An advanced open-source simulation tool designed for marine robotics, with a ros interface,'' in \emph{Proc. IEEE OCEANS Conf.}\hskip 1em plus 0.5em minus 0.4em\relax IEEE, 2019, pp. 1--6.

\bibitem{munoz2024learning}
B.~Mu{\~n}oz and G.~Troni, ``Learning the ego-motion of an underwater imaging sonar: A comparative experimental evaluation of novel cnn and rcnn approaches,'' \emph{IEEE Robot. Automat. Lett.}, 2024.

\bibitem{wang2023neus2}
Y.~Wang, Q.~Han, M.~Habermann, K.~Daniilidis, C.~Theobalt, and L.~Liu, ``{NeuS2}: Fast learning of neural implicit surfaces for multi-view reconstruction,'' in \emph{Proc. IEEE/CVF Int. Conf. Comput. Vis.}, 2023, pp. 3295--3306.

\bibitem{kerbl20233d}
B.~Kerbl, G.~Kopanas, T.~Leimk{\"u}hler, and G.~Drettakis, ``{3D} {Gaussian} splatting for real-time radiance field rendering,'' \emph{ACM Trans. Graph.}, vol.~42, no.~4, 2023.

\bibitem{Matsuki:Murai:etal:CVPR2024}
H.~Matsuki, R.~Murai, P.~H.~J. Kelly, and A.~J. Davison, ``{G}aussian splatting {SLAM},'' in \emph{Proc. IEEE Conf. Comput. Vis. Pattern Recognit.}, 2024.

\end{thebibliography}
\end{document}